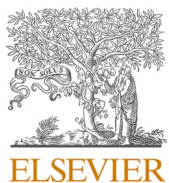
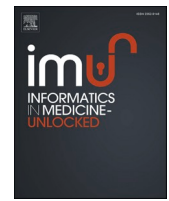

# CovidExpert: A Triplet Siamese Neural Network framework for the detection of COVID-19

Tareque Rahman Ornob [*], Gourab Roy, Enamul Hassan

*Department of Computer Science & Engineering, Shahjalal University of Science and Technology, University Ave, Sylhet, 3114, Bangladesh*

A B S T R A C T

Patients with the COVID-19 infection may have pneumonia-like symptoms as well as respiratory problems which may harm the lungs. From medical images, coronavirus illness may be accurately identified and predicted using a variety of machine learning methods. Most of the published machine learning methods may need extensive hyperparameter adjustment and are unsuitable for small datasets. By leveraging the data in a comparatively small dataset, few-shot learning algorithms aim to reduce the requirement of large datasets. This inspired us to develop a few-shot learning model for early detection of COVID-19 to reduce the post-effect of this dangerous disease. The proposed architecture combines few-shot learning with an ensemble of pre-trained convolutional neural networks to extract feature vectors from CT scan images for similarity learning. The proposed Triplet Siamese Network as the few-shot learning model classified CT scan images into Normal, COVID-19, and Community-Acquired Pneumonia. The suggested model achieved an overall accuracy of 98.719%, a specificity of 99.36%, a sensitivity of 98.72%, and a ROC score of 99.9% with only 200 CT scans per category for training data.

## 1. Introduction

The 2019 coronavirus disease (COVID-19) has now infected humans, causing severe acute respiratory illness. Since its first discovery in Wuhan, China, in December 2019, it has swiftly expanded worldwide. The World Health Organization (WHO) named the new epidemic illness Coronavirus Disease (COVID-19) after the virus that causes severe acute respiratory syndrome (SARS-CoV-2). And over 661 million COVID-19 symptoms have been confirmed as of December 26, 2022, and more than 6.685 million fatalities globally [1].

Due to a lack of critical care facilities and an overwhelming number of patients, health systems have collapsed, even in industrialised nations. On January 30, 2020, the WHO deemed the COVID-19 epidemic a worldwide health emergency; this was considered a pandemic on March 11, 2020 [2]. The symptoms of Coronavirus evolve within 2–14 days after viral infection.

Due to COVID-19's high infection rates, early coronavirus diagnosis is essential for treating the disease. To diagnose COVID-19, the actual-time reverse-transcription polymerase chain reaction (RT-PCR) analysis is now the standard method. RT-PCR is still the most accurate method for making a confident detection of COVID-19 infection [3]. Even though the absence of an effective RT-PCR assay, the high false detection rate with the protracted turnaround time (minimum 4–6 hours) makes it difficult to identify infected people quickly. Because of this, many infected people go undetected and often spread the infection to others. The incidence of COVID-19 illness should decrease with the early diagnosis of this condition. The inefficiency and scarcity of the current COVID-19 tests have motivated several initiatives to investigate alternative test methodologies. One method uses radiological imaging to identify COVID-19 infections, including X-rays or computed tomography (CT). The researchers believe that a technique based on chest CT scans might be a vital tool for identifying and quantifying COVID-19 instances [4].

CT scan diagnosis may benefit significantly from using artificial intelligence (AI) based techniques. Researchers are making global efforts to harness AI as a formidable tool to develop quick and affordable diagnostic methods to stop the current pandemic [5]. The main objectives of the study are the transmission of COVID-19, early detection, the creation of efficient treatments, and comprehension of its economic impacts.

Compared to standard head X-rays, CT scans include more precise information about tissue and structure, enabling the diagnosis of more traumas and disorders [6]. X-ray and CT scans are the two imaging methods most frequently used to examine COVID-19 patients. In this study, we concentrated on CT scans because chest CT and ultrasound of the lungs have been found to be more sensitive and moderately specific than chest X-Ray (CXR) in diagnosing COVID-19 [7]. Although deep learning methods have advanced significantly in various medical areas, the problem of dependency on the vast quantity of data still limits them. Researchers used a variety of strategies, including data augmentation and generative adversarial networks (GAN), to address this issue of data shortage [8]. Different data augmentation strategies have over-fitting issues, while GAN image generation techniques have difficulties simulating actual patient data, which results in unintended bias during experimental validation [8]. Additionally, several research studies have used various pre-trained Convolutional Neural Network (CNN) models in transfer learning approaches.

Given that these CNN models were pre-trained on a big non-medical dataset (i.e., ImageNet), significant fine-tuning is required to get promising diagnostic findings which often involves a lengthier training time.

To overcome these limitations, researchers are actively investigating few-shot learning (FSL) which is an instantiation of meta-learning. Meta-learning, also known as "learning to learn," is a machine learning technique that involves learning a learning algorithm that can adapt quickly to new tasks or environments [9]. In other words, meta-learning involves learning how to learn. Meta-learning can be used to learn a learning algorithm that can quickly adapt to new tasks or datasets, without the need for a large amount of labelled data for each task. This can be useful in situations where it is difficult or expensive to obtain a large amount of labelled data for a particular task. There are a number of ways to apply meta-learning to machine-learning tasks, depending on the specific problem the users are trying to solve. Some common approaches to meta-learning include FSL, transfer learning, and multi-task learning.

Few-Shot Learning is a form of meta-learning when a learner is provided experience on a range of related tasks to effectively transition to new (but related) activity with a constrained number of situations in the meta-testing stage [10]. In this process, the model is trained on a small number of examples or "shots" within each class in the training set. In few-shot learning, the goal is to learn a model that can adapt quickly to new tasks or datasets by using a small number of labelled examples. One of the main motivations for using few-shot learning is the need to reduce the data required for training a machine learning model. Since datasets are difficult to find (such as cases of a rare illness) or the cost of annotating data is high, few-shot learning is advantageous. Few-shot learning allows the model to learn from a small number of examples, which can be much easier and faster to obtain. The model does not need to identify the images in the training set while going on to the test set to achieve the aim of few-shot learning. Instead, the goal is to compare and contrast items' similarities and differences [11]. Given the rarity of big, well-balanced medical image datasets, this concept might be instrumental in image analysis. Overall, few-shot learning is a useful technique for reducing the data required for training machine learning models, and it has shown promising results in a range of applications.

The Siamese network is one of the meta-learning models which has lately been successful in applying FSL in various fields. Siamese networks are a particular network architecture used to create and evaluate feature vectors for each input. They comprise two or more similar (i.e., configured using the same parameters and weights) subnetworks [12]. By comparing the feature vectors of the Siamese network and training a similarity function between labelled points, it is possible to determine how similar the inputs are. To categorize COVID-19 patients with few training CT images, this research utilizes the Triplet Siamese Network, which consists of three identical subnetworks. To create feature embeddings from the input images, we employed an ensemble of pre-trained CNNs that had been fine-tuned, and we used the pairwise margin ranking loss function to change the network weights. Ensembling combines many models to create a solid and trustworthy model for making predictions.

The contributions of this study are outlined in the following.

1. Developing an ensemble of pre-trained CNN, which utilized the Triplet Siamese network to assist in the early diagnosis of patients with COVID-19.
2. Utilizing a CT scan dataset with 600 training images to identify COVID-19.
3. The advantage of employing Siamese networks with small datasets is the main emphasis of the suggested study.
4. A performance assessment is provided to show how effective the suggested framework is.

The paper is assembled as follows: An overview of recent literature articles related to this study is given in Section 2. Section 3 describes the proposed system, along with gathering and preparing datasets. Section 4 provides the experimental outcomes, comparative performance of the suggested framework and discussion. The study is concluded with further research in Section 5.

## 2. Related works

Researchers have created deep learning approaches for detecting COVID-19 utilizing medical images, CXR, and CT scans to prevent the COVID-19 epidemic. This review discusses the most current COVID-19 detection systems that have used deep learning methods.

To expedite the identification of COVID-19 victims, Panwar et al. [13] suggested a VGG-19 transfer learning method that utilizes CT-Scan and chests X-ray data. They utilized the COVID-19 chest X-ray dataset that includes 673 radiological images from 342 individuals. And the SARS-COV-2 CT scan dataset contains 1230 CT scans of those who confirmed negative for COVID-19 and 1252 CT images of individuals who confirmed positive for COVID-19. Their proposed model has a 95.61% COVID-19 case identification accuracy. According to their methodology, a patient identified with pneumonia is more likely to test as a False Positive.

A completely automated technique to recognize COVID-19 illness from the chest CT results of the patients was proposed by Rahimzadeh et al. [14]. The ResNet50V2 model served as the framework for their experiment. Their dataset included 48,260 CT scans comprising 282 healthy people and 15,589 images comprising 95 patients with COVID-19. The suggested model was performed on 7996 test images at this binary image classification task with 98.49% accuracy.

Utilizing chest X-ray and CT scan images, Hussain et al. [15] created a novel CNN model (CoroDet) to identify COVID-19 automatically. For the dataset, they utilized 7390 images. The proposed model produced a classification accuracy of 99.1% for two-class classification (COVID and Normal), 94.2% for three-class classification (COVID, Normal, and non-COVID pneumonia), and 91.2% for four-class classification (COVID, Normal, non-COVID viral pneumonia, and non-COVID bacterial pneumonia). The experiment shows 500 and 800 regular images for each category of two, three, and four classes, respectively. There are 400 images for the 4-class categorization of viral and bacterial pneumonia. For the three classes, there are 800 images of bacteria that cause pneumonia. The doctor claims that their suggested model is not ready to take the place of current COVID-19 detection tests.

To identify COVID-19, Kogilavani et al. [16] used CNN architectures such as VGG16, DeseNet121, MobileNet, NASNet, Xception, and EfficientNet. The collection contains 3873 overall CT scan images, including COVID-19 and Non-COVID scans. The accuracy rates are 97.68%, 97.53%, 96.38%, 89.51%, 92.47%, and 80.19% for VGG16, DenseNet121, MobileNet, NASNet, Xception, and EfficientNet, respectively. VGG16 design offered more accuracy in their analysis compared to other architectures.

CNN with KNN (K-Nearest Neighbour) was used by Basu et al. [17] to provide a comprehensive structure for detecting COVID-19 using CT scan images. They used the SARS-COV-2 CT-Scan Dataset, which was upgraded from the original 2482 CT scan data to contain 2926 images. On their dataset, their model produced classification accuracy for binary classification of 97.30% and 98.87%, respectively. The disadvantage of the suggested technique is that it cannot identify CT scans that are COVID-19 positive during the initial phases of the infection.

An approach to identify COVID-19 using the VGG-19 transfer learning model was suggested by Horry et al. [18]. They utilized 1103 images for the ultrasound image dataset (COVID-19, Pneumonia, and Normal), 60798 images for the X-ray dataset (COVID-19, Pneumonia, and Normal), and 746 images for the CT scan dataset (COVID and Non-COVID). Their tuned VGG19 model with the right settings performs at substantial levels of COVID-19 detection versus pneumonia or normal for all three lung imaging methods, with an accuracy of up to 86% for X-rays, 100% for ultrasounds, and 84% for CT scans.

A BiT-M classification algorithm to detect COVID-19 using CT scan images was suggested by Zhao et al. [19]. They utilized the COVIDx CT-2A dataset, which contains 194,992 CT scan images. According to the study, the suggested multi-class model had a 99.2% accuracy rate for identifying COVID-19 instances. Their proposed model is at a theoretical research stage, and the model hasn't been put to the test in actual clinical practice.

Serte et al. [20] published a model to forecast COVID-19 on a 3D CT volume with ResNet-50 and majority voting. They used the 3D CT scan images from the Mosmed and CCAP datasets. They transformed the 3D CT images into 2D images for training and assessment. 5019 images were utilized for testing, while 5493 images were used for training. The suggested deep learning model has an AUC value of 96% for identifying COVID-19 on CT images. Their model may not function on a phone or web server since they don't have the high memory requirements of modern desktop or laptop computers.

Mukherjee et al. [21] suggested a nine-layer Convolutional Neural Network that can simultaneously train and evaluate CXRs and CT scans. They used a total of 672 images in the dataset, of which 336 were chest X-rays and 336 were CT scans. Their suggested model attained an overall accuracy of 96.28% in a test dataset consisting of 135 images.

A model known as CTnet-10 that had been created for the COVID-19 diagnosis was suggested by Shah et al. [22]. They used a dataset that had 738 images overall. Their model has an 82.1% accuracy rate. Compared to other models, the accuracy of their proposed model is quite low.

An ensemble deep learning model using pre-trained Residual Attention and DenseNet architectures was developed by Maftouni et al. [23]; and it performed as a reliable COVID-19 classifier on noisy labelled chest CT scan images. Their dataset included 2618 CAP images representing 60 cases, 6893 normal images representing 604 cases, and 7593 COVID-19 images representing 466 cases. On the test set, their ensemble model achieved 95.31% accuracy.

Shorfuzzaman et al. [24] created a synergistic strategy to integrate contrastive learning with a fine-tuned ConvNet encoder to get unbiased feature representations and employed a Siamese network for the final classification of COVID-19 cases. They created a balanced dataset for this investigation with 678 Chest X-ray images. They pre-trained the VGG-16 encoder network with 480 images during training and 198 images in testing. 30 images were used for the Siamese network's training and 648 images for its testing. The accuracy of their suggested model was 95.6%.

Jadon et al. [25] suggested employing Siamese networks with a few-shot learning method to find COVID-19. A total of 4200 chest X-ray images from 3 classes—COVID-19, Viral Pneumonia, and Normal—were included in the study's dataset. Their recommended method allowed them to attain 96.4% accuracy.

Most of the strategies used in the literature to diagnose COVID-19 using CT or CXR images either employed pre-trained transfer learning models or custom CNN architecture, both of which need a lot of training data. Although some research employed a small amount of training data, their test dataset was so small that it is questionable if their model could help physicians in a real-world situation. As an alternative, we have suggested a Triplet Siamese Network that uses an ensemble of pre-trained BiT, DenseNet121, SwinTransformer, MobileNetV2, EfficientNetB0, and ResNext as the encoder of the architecture to classify COVID-19 instances with just 600 training CT images. To ensure the model worked in a real-world situation, we tested it with a dataset of 10152 images. Our proposed model is computationally efficient that can achieve better or the same level of accuracy as the pre-trained and other custom CNN models. Additionally, categorical cross entropy or binary cross entropy loss function is the most used strategy in the literature. Contrarily, we used the margin ranking loss function, leading to a quicker model adaptation rate with fewer tests and updates to the hyperparameters. Additionally, most available models involve image augmentation to increase model generalizability, which lengthens training time. The proposed few-shot learning model performs better than previous studies' models despite not using augmentation.

## 3. Methodology and dataset

The COVID-19 detection system, which consists of many stages, is shown in Fig. 1. The preprocessing pipeline was initially applied to the raw CT scan images. Data scaling, shuffles, noise removal, image sharpening, normalizing, brightness and contrast adjustments were made in the pre-processing pipeline. The preprocessed data set was then partitioned into two sets of training dataset and testing dataset, where one set was used to train transfer-learning-based models, and another set was used for the proposed Triplet Siamese Network. After each epoch, the training accuracy and loss were computed. Validation accuracy and loss were simultaneously found using 10-fold cross-validation. The confusion matrix, accuracy, ROC AUC, specificity, sensitivity, and F1-score were utilized to test the performance of the suggested system.

### 3.1. Dataset description

The dataset is taken from Maftouni et al. [23]. Based on the authors, this COVID-19 lung CT dataset is the biggest one. This dataset contains 2618 images of community-acquired pneumonia, 7593 COVID-19 images, and 6893 normal images. By combining CT scan data from seven open datasets, they created the COVID-19 dataset. These datasets have shown their effectiveness in deep learning applications by being widely utilized in the COVID-19 diagnostic literature. The combined dataset is thus expected to boost the classification performance of deep learning techniques by merging all of these resources. Table 1 displays the dataset's properties.

From the author-provided source, we created two datasets. One is for CNN models based on transfer learning, whereas the other is for the FSL model. We employed a total of 12344 CT scans for training and 3390 CT scans for evaluating the transfer-learning-based CNNs. The dataset distribution for transfer-learning-based models is shown in Table 2.

We created a meta-dataset for the FSL model that includes images of the Anchor (an arbitrary class of points), Positive (the same class as the Anchor), and Negative (a different class from the Anchor). We employed 10152 CT scans for the test dataset, whereas 600 CT images were used to train this model. A radiologist helped us create the meta-dataset by arranging the slices of the CT scans so that the anchor and positive slices are identical and the anchor and negative slices are different. Since the two CT scans must be clinically related and not only belong to the same class, randomly producing an Anchor and Positive pair would've been incorrect. The selected balanced dataset for the suggested Triplet Siamese Neural Network is shown in Table 3.

Fig. 2 and Fig. 3 shows input image for the models.

### 3.2. Data pre-processing

There was a visible difference among CT scan slices taken at the exact location on the chest because the CT scans were obtained from clinics in several different nations. The majority of the images featured noise, and many of them were blurry (i.e., bed rest corner, etc.). Since the data is improved after noise reduction, CNN can capture essential features of the images while ignoring the noises, making noise removal crucial. Background removal, automatic brightness and contrast adjusting, and unsharp masking were used to fix these problems.

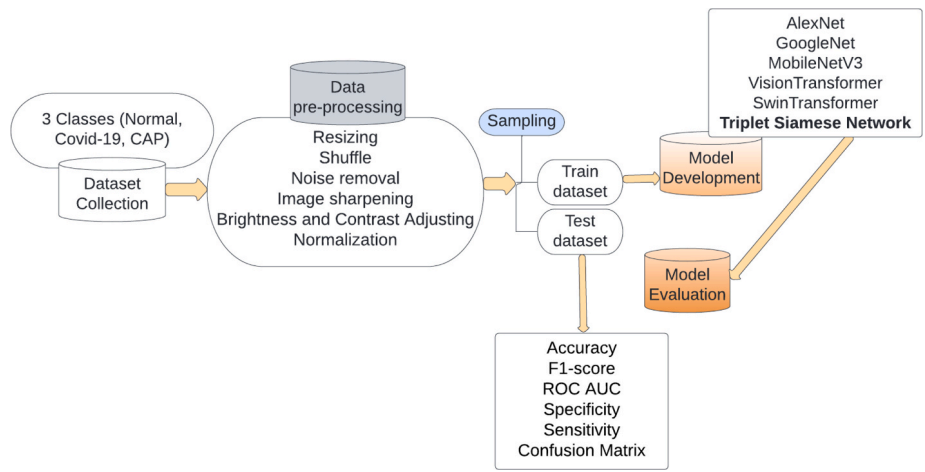

**Fig. 1.** Pipeline to develop the proposed model.

**Table 1**
Dataset statistics.

| Case | Male | Female | Unknown Gender | Overall | Age |
|---|---|---|---|---|---|
| COVID-19 | 3781 | 2240 | 1572 | 7593 | $51.26 \pm 16.49$ |
| Normal | 4244 | 2649 | 0 | 6893 | $52.82 \pm 21.87$ |
| CAP | 1567 | 1051 | 0 | 2618 | $58.14 \pm 20.94$ |
| Overall | 9592 | 5940 | 1572 | 17104 | $53.10 \pm 19.92$ |

**Table 2**
Dataset Distribution for transfer-learning-based models.

| Dataset | COVID-19 | Normal | CAP | Total |
|---|---|---|---|---|
| Training | 5479 | 4968 | 1897 | 12344 |
| Validation | 608 | 552 | 210 | 1370 |
| Testing | 1506 | 1373 | 511 | 3390 |
| Overall | 7593 | 6893 | 2618 | 17104 |

**Table 3**
Dataset Distribution for the proposed framework.

| Dataset | COVID-19 | Normal | CAP | Total |
|---|---|---|---|---|
| Training | 200 | 200 | 200 | 600 |
| Validation | 20 | 20 | 20 | 60 |
| Testing | 3384 | 3384 | 3384 | 10152 |
| Overall | 3604 | 3604 | 3604 | 10812 |

*3.2.1. Unsharp masking*
A linear image processing method that sharpens the image is called unsharp masking. The sharp details may show the difference between the original image and its blurred counterpart. The original image is then resized to include these elements:

$$Enhanced image = original + amount * (original - blurred)$$

Any image filter technique could be used for the blurring stage; however, we used the gaussian filter. The particular parameters of the gaussian filter size and the weights when the images are subtracted define the exact attributes of the filter. The unsharp mask increases the high-frequency components of the image.

*3.2.2. Background removal*
A function was implemented to normalize the pixel values to eliminate the background and the bed rest. We first determined the average pixel value near the foreground to renormalize washed-out images. Then, KMeans was used to establish the threshold for differentiating between the foreground (soft tissue/bone) and background (lung/air). To avoid accidently clipping the lung, we erode the smaller components and expand to include some of the pixels around it. The biggest contour in the intermediate image was located after the initial foreground mask was discovered. After initializing a second foreground mask, a bitwise AND operation of the two images was performed to create an intersection of the two foreground masks. After filling the holes in the enclosed mask, we applied the mask to the image. We removed the background and the associated noises by cropping the masked image.

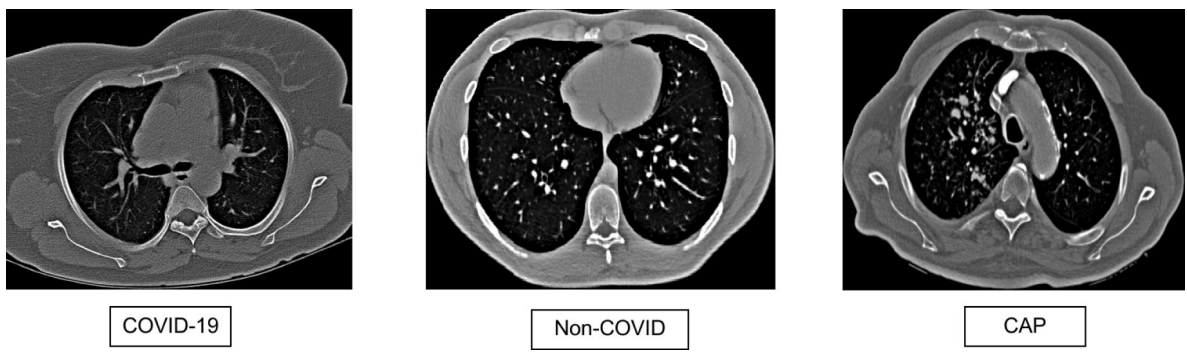

**Fig. 2.** Input Data Visualization for transfer-learning-based CNN models.

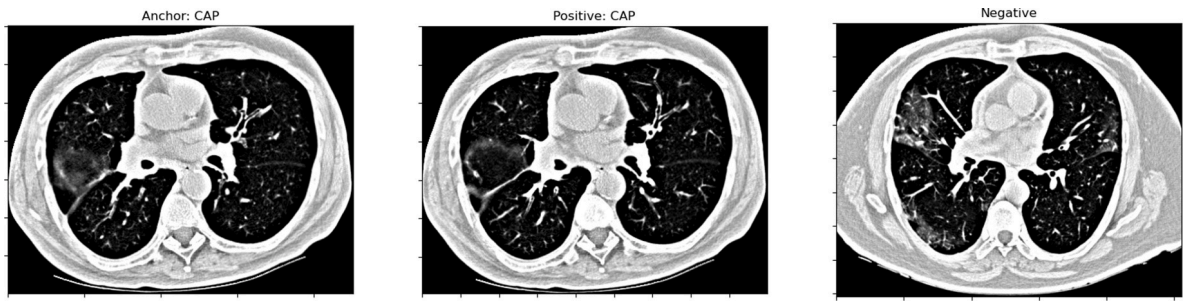

**Fig. 3.** Input Data Visualization for the proposed framework.

### 3.2.3. Automatic brightness and contrast adjusting

The brightness and contrast may be modified using alpha ($\alpha$) and beta ($\beta$). These are sometimes referred to as the gain and bias parameters. The expression is written as follows:

$$g(x) = \alpha * f(x) + \beta$$

where the pixels for the source image are f(x) and the pixels for the output image are g(x). Conveniently, the phrase may be written as:

$$g(i,j) = \alpha * f(i,j) + \beta$$

where i and j denote the pixel's location as being in the i-th row and j-th column.

This was accomplished by examining the image's histogram. $\alpha$ and $\beta$ are automatically calculated for [0 … 255] output range for brightness and contrast improvement. When the color frequency fell below a certain threshold, we computed the cumulative distribution to decide where to cut the right and left sides of the histogram. As a result, we now have our minimum and maximum ranges. Our target output spectrum of 255 was divided by the lowest and highest grayscale range after clipping to determine $\alpha$.

$$\alpha = 255 / (maximum_{gray} - minimum_{gray})$$

We used the formula $g(i,j) = 0$ and $f(i,j) = minimum_{gray}$ to calculate $\beta$, and the result was $g(i,j) = \alpha * f(i,j) + \beta$ which yields $\beta = -minimum_{gray} * \alpha$.

Thus, by solving $\alpha$ and $\beta$ introducing $\alpha$ and $\beta$ to the images using saturation arithmetics, we were able to improve brightness and contrast. Fig. 4 shows us the effect of data pre-processing on the original dataset.

### 3.3. Development of transfer-learning-based models

The machine learning method of transfer learning uses a model developed for one job as the foundation for another. It is a typical deep learning technique to use pre-trained models as the foundation for computer vision and natural language processing tasks due to the massive computing and time assets required to construct neural network algorithms for these problems and the huge jumps in the skill that they provide on related issues. It is common to perform transfer learning with predictive modelling problems that use image data as input. For these cases, it is customary to use a deep learning model that has been pre-trained for a big and challenging image classification task, like the ImageNet [26]. classification competition. Numerous models have been trained using the ImageNet dataset. Due to their cutting-edge architecture, we used the AlexNet, GoogLeNet, MobileNetV3, Vision Transformer, and Swin Transformer models to categorize COVID-19 using the initial dataset. Each of these models was trained using a simple augmentation technique (i.e., RandomRotation, RandomVerticalFlip, RandomHorizontalFlip, RandomAffine, ColorJitter, RandomAdjustSharpness, RandomAutocontrast) for 10 epochs that used the Adam optimizer with a batch size of 64 for AlexNet, 32 for GoogLeNet and MobileNetV3, and 16 for VisionTransformer and SwinTransformer. 0.0001 was the learning rate for every model except VisionTransformer, and the learning rate for VisionTransformer was 0.00001. The loss function in these models was Focal loss.

### 3.3.1. AlexNet

One of the first CNN proposed by Krizhevsky et al. [27] was AlexNet. There are eight weighted layers in the AlexNet (i.e., five convolutional layers and three fully connected layers).

### 3.3.2. GoogLeNet

GoogLeNet [28] is an image classification and recognition deep learning convolution neural network architecture. The GoogleNet architecture consists of 22 layers and a total of 9 inception modules (when pooling layers are included, there are 27 layers). GoogLeNet employs nine initial layers because adding more layers and more data is the most effective strategy to increase the output of deep learning models.

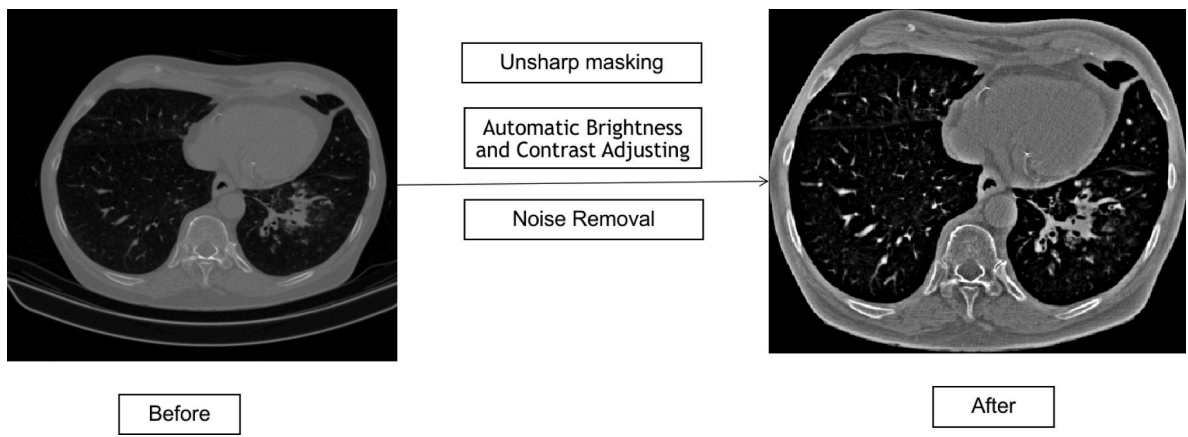

**Fig. 4.** Comparison between the original and pre-processed image.

### 3.3.3. MobileNetV3

Hardware-aware network architecture search (NAS) and Platform-Aware Neural Network Adaptation for Mobile Applications (NetAdapt) are the foundations of MobileNetV3 [29]. To facilitate calculation, MobileNetV3 has included the Squeeze and Excitation (SE) module, and the sigmoid in the SE module has been replaced with a Hard-Sigmoid. ReLU is replaced with Swish to increase non-linearity.

### 3.3.4. Vision transformer

Without convolutional layers, the Vision Transformer (ViT) [30] is a modified NLP Transformer (encoder alone) for image categorization. Patches of an image are separated, flattened, and then placed in a two-dimensional embedding space. Each vector receives an additional token to indicate its relative position inside the image, and the whole sequence of vectors receives an additional, learnable token to indicate the class. The vector sequence is put into the conventional Transformer encoder, which has been altered to include an additional fully-connected layer for classification at the very end.

### 3.3.5. SwinTransformer

A division of Vision Transformers are Swin Transformers (ST). Due to the self-attention processing happening only inside each local window, it has a linear computational cost proportionate to the size of the input image. It builds hierarchical feature maps by combining image patches into deeper layers [31]. In contrast, owing to global self-attention processing, early vision transformers build feature maps with a single low resolution and have a quadratic computational cost proportionate to the size of the input image.

### 3.3.6. ResNetV2-101

Modern transfer learning techniques for image categorization include BigTransfer (BiT) [32]. When training deep neural networks for vision, the transfer of previously learned representations enhances sampling efficiency and makes hyperparameter tweaking easier. Pre-training on large supervised datasets and model fine-tuning on a target task are paradigms that BiT revisits. ReseNetV2-101 is a BiT model that is 101 layers deep.

### 3.3.7. DenseNet121

The first layer of the DenseNet [33] convolutional neural network is linked to the second, third, fourth, and so on, while the second layer is connected to the third, fourth, fifth, and so on layer. Each layer of the network is interconnected with all layers that are deeper in the network. Apart from the fundamental convolutional and pooling layers, DenseNet is made up of two significant building elements. They consist of Transition layers and Dense Blocks. The four dense blocks in the DenseNet-121 have [6,12,24,16] layers.

### 3.3.8. MobileNetV2

In contrast to standard residual models that employ extended representations in the input, the MobileNetV2 [34] architecture is built on an inverted residual structure where the input and output of the residual block are narrow bottleneck layers. Lightweight depthwise convolutions from MobileNetV2 filter features in the intermediate expansion layer. To retain representational strength, non-linearities in the thin layers were also eliminated.

### 3.3.9. EfficientNet-B0

Using the ImageNet dataset as its training data, EfficientNet-B0 [35] is a simple mobile-size baseline architecture with 237 layers. The model uses seven inverted residual blocks, however, they each have unique settings. In addition to swish activation, these blocks also use squeeze and excitation blocks.

### 3.3.10. ResNeXt-101

The ResNeXt101 [36] is a CNN model that is based on a conventional ResNet model but uses 3x3 convolutions instead of 3x3 grouped convolutions within the bottleneck block. One convolution is divided into many smaller, parallel convolutions by the ResNeXt bottleneck block. The cardinality of the ResNeXt101 model is 32, and the bottleneck width is 4. This implies that 32 parallel convolutions with just 4 filters are employed in place of a single convolution with 64 filters.

### 3.3.11. Focal loss

A balanced dataset for multi-class classification includes equally distributed target labels. An unbalanced dataset is one where one class contains significantly more samples than the other. This mismatch results in two issues.

- Training is ineffective since the majority of examples are simple ones that don't provide any relevant learning signals.
- The abundance of simple instances during training may degrade models.

Focal Loss (FL) [37] is a modified form of Cross-Entropy Loss (CE) that attempts to address the class imbalance problem by giving more weight to complex or easily misclassified cases and less weight to easy examples. As a result, Focal Loss decreases the loss contribution from simple examples while emphasizing the need to address misclassified cases. Due to the primary dataset's extreme imbalance, we chose to employ FL rather than CE since it produced superior results.

Mathematically, the focal loss is defined as:

$$FL(p_t) = -\alpha_t (1-p_t)^\gamma \log(p_t)$$

where the likelihood of the ground truth label, denoted by $p_t$, might be expressed as follows:

$$p_t = \begin{cases} p & \text{if } y = 1 \\ 1-p & \text{otherwise} \end{cases}$$

For Focal Loss, there are two adjustable parameters. The focusing parameter $\gamma$ gradually adjusts the rate at which simple instances are down-weighted. When $\gamma = 0$, FL equals CE, and as $\gamma$ increases, the modulating factor's influence also grows ($\gamma = 2$ was the most effective in the studies). Focal loss is balanced by $\alpha$, and the accuracy was marginally higher than the non-$\alpha$-balanced variant.

### 3.4. Algorithmic details of the proposed system

The Triplet Siamese Network, a Siamese network with three identical subnetworks, is used in this research. We fed the model three images, two of which were similar (anchor and positive samples) while the third was dissimilar (a negative example). A representation of this concept has been shown in Fig. 5.

Our goal was for the model to learn to estimate the similarity between images. The Siamese network received each triplet image as an input, generated the embeddings, and output the distance between the anchor and the positive embedding and the distance between the anchor and the negative embedding.

Six transfer-learning based models were used as the backbone of the Triplet Siamese Network to create an ensemble model to generate embeddings for each of the images of the input triplet images. The ensemble model included ResNetV2, DenseNet, SwinTransformer, MobileNetV2, EfficientNetB0, ResNeXt-101. The final layer of each model consisted of 512 neurons. The weights of all the layers except the final layer of the six models were frozen. This is important to avoid affecting the weights that the model has already learned. The final layer was left trainable so that we could fine-tune its weight during training.

Every model generated embeddings independently, and the concatenation of embeddings was fed into a neural network of 512 neurons to reduce the generalization error of the extraction. Fig. 6 illustrates the proposed model architecture.

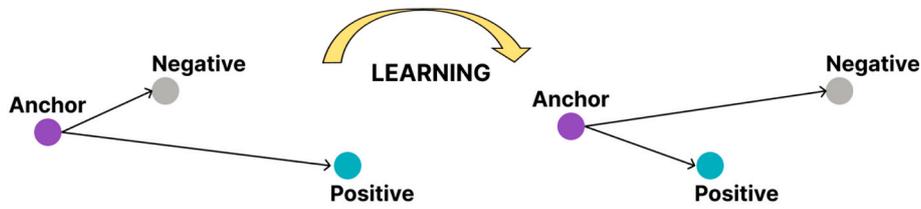

**Fig. 5.** Triplet input (anchor, positive and negative).

The PairwiseDistance method was employed to compute the pairwise distances between the Anchor image's vectors and the Positive image's vectors, and the Anchor image's vectors and the Negative image's vectors using the p-norm. The pairwise distances were to measure the similarity between those vectors. The formula for the PairwiseDistance method is:

$$||x||_p = \left( \sum_{i=1}^{n} |x_i|^p \right)^{1/p}$$

Marking Ranking Loss was used as the loss function. The goal of Marking Ranking Loss is to anticipate the relative distances between inputs, as opposed to other loss functions like Cross-Entropy Loss or Mean Square Error Loss, whose goal is to learn to predict directly a label,

| Layer (type:depth-idx) | Output Shape | Param # |
|---|---|---|
| Ensemble | [1, 512] | -- |
| ├─Bit: 1-1 | [1, 512] | -- |
| │    └─ResNetV2: 2-1 | [1, 512] | -- |
| │         └─Sequential: 3-1 | [1, 64, 56, 56] | (9,408) |
| │         └─Sequential: 3-2 | [1, 2048, 7, 7] | (42,478,976) |
| │         └─GroupNormAct: 3-3 | [1, 2048, 7, 7] | (4,096) |
| │         └─ClassifierHead: 3-4 | [1, 512] | 1,049,088 |
| ├─DenseNet121: 1-2 | [1, 512] | -- |
| │    └─DenseNet: 2-2 | [1, 512] | -- |
| │         └─Sequential: 3-5 | [1, 1024, 7, 7] | (6,953,856) |
| │         └─Linear: 3-6 | [1, 512] | 524,800 |
| ├─SwinTransformer: 1-3 | [1, 512] | -- |
| │    └─SwinTransformer: 2-3 | [1, 512] | -- |
| │         └─PatchEmbed: 3-7 | [1, 3136, 128] | (6,528) |
| │         └─Dropout: 3-8 | [1, 3136, 128] | -- |
| │         └─Sequential: 3-9 | [1, 49, 1024] | (86,734,648) |
| │         └─LayerNorm: 3-10 | [1, 49, 1024] | (2,048) |
| │         └─AdaptiveAvgPool1d: 3-11 | [1, 1024, 1] | -- |
| │         └─Linear: 3-12 | [1, 512] | 524,800 |
| ├─MobileNetV2: 1-4 | [1, 512] | -- |
| │    └─MobileNetV2: 2-4 | [1, 512] | -- |
| │         └─Sequential: 3-13 | [1, 1280, 7, 7] | (2,223,872) |
| │         └─Sequential: 3-14 | [1, 512] | 655,872 |
| ├─EfficientNetB0: 1-5 | [1, 512] | -- |
| │    └─EfficientNet: 2-5 | [1, 512] | -- |
| │         └─Sequential: 3-15 | [1, 1280, 7, 7] | (4,007,548) |
| │         └─AdaptiveAvgPool2d: 3-16 | [1, 1280, 1, 1] | -- |
| │         └─Sequential: 3-17 | [1, 512] | 655,872 |
| ├─Resnext101_32×8d: 1-6 | [1, 512] | -- |
| │    └─ResNet: 2-6 | [1, 512] | -- |
| │         └─Conv2d: 3-18 | [1, 64, 112, 112] | (9,408) |
| │         └─BatchNorm2d: 3-19 | [1, 64, 112, 112] | (128) |
| │         └─ReLU: 3-20 | [1, 64, 112, 112] | -- |
| │         └─MaxPool2d: 3-21 | [1, 64, 56, 56] | -- |
| │         └─Sequential: 3-22 | [1, 256, 56, 56] | (420,864) |
| │         └─Sequential: 3-23 | [1, 512, 28, 28] | (2,405,376) |
| │         └─Sequential: 3-24 | [1, 1024, 14, 14] | (55,160,832) |
| │         └─Sequential: 3-25 | [1, 2048, 7, 7] | (28,745,728) |
| │         └─AdaptiveAvgPool2d: 3-26 | [1, 2048, 1, 1] | -- |
| │         └─Linear: 3-27 | [1, 512] | 1,049,088 |
| ├─Linear: 1-7 | [1, 512] | 1,573,376 |

Total params: 235,196,212
Trainable params: 6,032,896
Non-trainable params: 229,163,316

**Fig. 6.** Architecture of the proposed model.

a value, or a group of values given an input. The formula for Margin Ranking Loss is:

$$loss(x, y) = max(0, -y * (x_1 - x_2) + margin)$$

A similarity score between the data points is necessary to employ this loss. The dissimilarity between the anchor image and the positive image has to be low, while the dissimilarity between the anchor image and the negative image has to be high. The ensemble model gathered features from the three input data points to create embedded representations for each so that the Margin Ranking Loss function could be applied. The PairwiseDistance method was used to get the similarity score.

As a result, the gradients were computed using this loss function, and the weights and biases of the siamese network were modified using the gradients. Finally, the feature extractors were trained to provide comparable representations for both inputs when the inputs are similar, or distant representations for the two inputs when they are dissimilar.

The high-level architecture of the proposed model to diagnose COVID-19, CAP, and Normal has been presented in Fig. 7. It includes the ensemble of models for feature embedding, similarity learning with Margin Ranking loss, and the training process.

### 3.5. Evaluation metrics

The performance of the proposed system was evaluated by using the eight metrics: Accuracy, Precision, Recall, F1-score, Specificity, and AUC ROC score. They were determined by:

$$Accuracy = \frac{TP + TN}{TP + FP + TN + FN}$$

$$Precision = \frac{TP}{TP + FP}$$

$$Recall = \frac{TP}{TP + FN}$$

$$F1 - score = \frac{2 * Precision * Recall}{Precision + Recall}$$

$$Specificity = \frac{TN}{TN + FP}$$

- TP stands for COVID-19 instances that were predicted properly
- FP stands for normal, or CAP instances that the proposed system incorrectly classifies as COVID-19

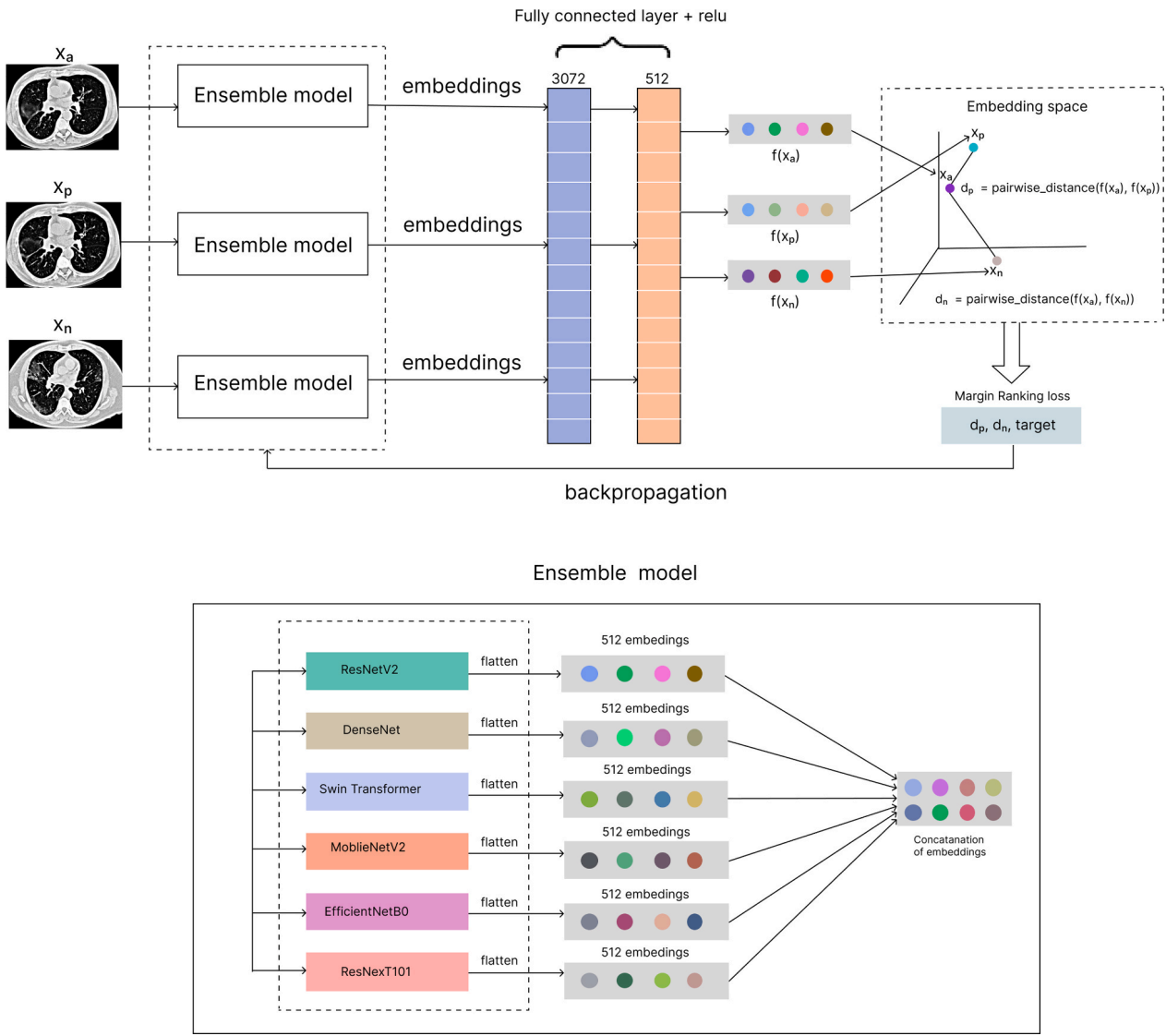

**Fig. 7.** An illustration of the proposed model.

- TN stands for the appropriately categorized normal or pneumonia cases
- FN stands for COVID-19 instances that were incorrectly categorized as pneumonia or normal cases.

## 4. Experimental result and discussion

### 4.1. Experimental environments

The models were developed on the AMD Ryzen Threadripper 1950X 16-Core Processor with 48 GB of main memory using Python 3.7.12 and Pytorch 1.11.0. The experiments were conducted utilizing two NVIDIA GeForce RTX 2080 Ti GPUs, each with 11 GB of RAM. The computer ran on Ubuntu 20.04.2 LTS.

### 4.2. Analysis and discussion of results

The transfer-learning-based models' dataset has been divided into 80% train dataset and 20% test dataset. The proposed Triplet Siamese Network dataset (i.e., meta-dataset) was divided into 600 images of the training dataset and 10152 images of the testing dataset. Since the dataset was partitioned in a patient-aware manner, no CT scan slice from the same patient was used in both the training and testing datasets. To get the results, the 10-fold cross-validation approach was utilized. To determine how well our proposed few-shot learning model would recognize COVID-19 occurrences, we first compare its accuracy and F1 score with several pre-trained CNN models. Table 4 provides the comparison results of the 3-class (Normal, CAP, COVID-19) classification.

Since the proposed model utilized ensembling learning, ablation studies on the models which were included in the ensemble model has been shown in Table 5.

Based on Table 5, we could conclude that it is beneficial to include all the six models in the ensemble as they provided complementary predictions. The base models made different types of errors on different examples and combining the predictions of these models resulted in a more diverse set of predictions and fewer overall errors, which led to improved performance. The proposed ensemble model smoothed out the predictions of the individual models, reduced the impact of overfitting and improved the generalization ability of the model.

As shown in Table 4, our suggested model outperforms the implemented pre-trained models with a substantially better score. The result is more significant because the proposed model's score was obtained from 10152 images, whereas the other models were tested with only 3390 images. Furthermore, compared to pre-trained CNN models, the suggested model's training and validation loss seem more stable and better convergent, as illustrated in Fig. 8. We have used the L2 dropout regularization strategy to prevent our model from overfitting to training data. Early stopping was employed with a learning schedule (the ReduceLROnPlateau callback from Pytorch) to prevent overfitting. Table 6 summarizes the classification performance results for additional verification of the performance of our proposed model on a class-by-class basis. A graphical representation of Table 6 has been given in Fig. 9.

The suggested model generates excellent values for Recall (98.719%) and Precision (98.719%), which are regarded as extremely important performance predictions for applications in medical contexts. Our model was trained using only 600 CT images, which is encouraging. Additionally, the suggested few-shot learning model consists of an ensemble of pre-trained CNN models, which increases the network's robustness and accuracy. Ensemble models are more trustworthy and robust than basic deep learning techniques. It unequivocally demonstrates the advantages of the few-shot learning strategy over other cutting-edge models for categorizing COVID-19 patients. Our method beat cutting-edge CNN models, with CovidExpert learning much better than Swin Transformer (Swin Transformer is the second best model according to F1 score); CovidExpert had a 98.719% accuracy, a specificity of 99.36%, a sensitivity of 98.719%, and an F1-score of 98.719%. For every performance indicator, the CovidExpert showed superior performance.

**Table 4**
Results from multiple models on the test dataset.

| Model | Accuracy | F1 score | Test Dataset |
|---|---|---|---|
| AlexNet | 0.71 | 0.71 | 3390 images |
| GoogleNet | 0.83 | 0.83 | |
| MobileNetV3-Large | 0.79 | 0.79 | |
| Swin Transformer | 0.85 | 0.85 | |
| **Proposed model (ours)** | **0.98719** | **0.98719** | **10152 images** |

**Table 5**
Ablation studies on the models used in the proposed ensemble model.

| Model | Accuracy | Precision | Recall | F1 score |
|---|---|---|---|---|
| ResNetV2 | 0.9564 | 0.9564 | 0.9564 | 0.9564 |
| DenseNet | 0.9714 | 0.9714 | 0.9714 | 0.9714 |
| Swin Transformer | 0.9706 | 0.9706 | 0.9706 | 0.9706 |
| MobileNetV2 | 0.9552 | 0.9552 | 0.9552 | 0.9552 |
| EfficientNetB0 | 0.9275 | 0.9275 | 0.9275 | 0.9275 |
| ResNeXT101 | 0.9708 | 0.9708 | 0.9708 | 0.9708 |
| **Proposed ensemble model (ours)** | **0.9872** | **0.9872** | **0.9872** | **0.9872** |

The confusion matrix from the model's inference stage is shown in Fig. 10. Only 28 COVID-19 images out of 10152 were incorrectly classified, which is better than human classification. As a result, COVID-19 instances may be classified effectively using the suggested approach.

Additionally, the ROC curves are shown in Fig. 11 to assess the overall efficacy by summarizing the trade-off between true positives and false positive rates. The suggested architecture's Receiver Operating Characteristic (ROC) curve was estimated to be 99.9%, demonstrating the CovidExpert's exceptional diagnostic capacity.

For all conceivable threshold values, the precision-recall (PR) curve in Fig. 12 displays the precision vs. the recall (true positive rate). High recall and high precision are desired outcomes. Similar trade-offs exist between high precision and high recall; when threshold is lowered, recall increases but precision decreases.

Table 7 compares our suggested model with the already published models.

It should be noted that the comparisons made in Table 7 are not based on the same datasets or data splits. Therefore, there is bound to be differences in the test score between the existing methods and the proposed model. But our test dataset was curated from the same dataset that Maftouni et al. [23] used.

According to Table 7, several of the binary classification methods in use today have marginally higher accuracy than our suggested model. But our suggested approach is better than other methods regarding multi-class classification. Only one author [19] evaluated their model for binary image classification using a testing dataset that was bigger than ours, but their training dataset included 194922 images as opposed to our training dataset's 600 images. We used a test dataset of 10152 images to evaluate the proposed model. This test dataset is substantially more extensive than the training dataset, and our dataset is diverse since it was collected from many nations. This phenomenon makes it clear that our methodology is excellent for helping physicians and radiologists.

A graphical user interface (GUI) was developed for model inference. Several images from the test dataset were predicted using the GUI. The model uses Cosine similarity to measure the similarity between the input image and a similar image from the query set. The model's inference on six images is shown in Figs. 13 and 14.

## 5. Conclusion

The number of COVID-19 cases is constantly rising, and many nations are experiencing resource difficulties. Therefore, it is essential to

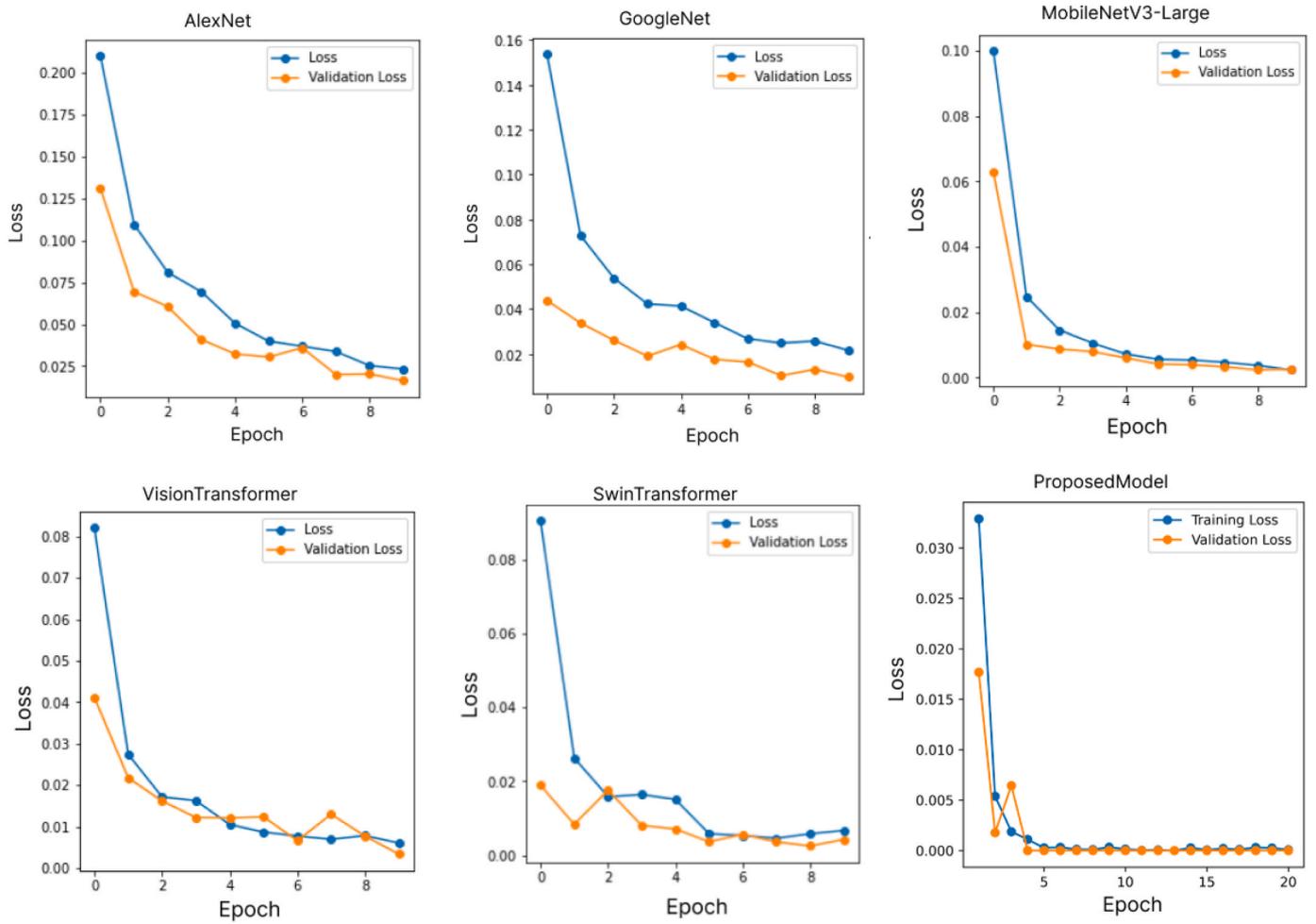

**Fig. 8.** Training vs. validation loss of pre-trained CNN models and the proposed model.

Table 6
The class-wise score of the proposed model.

|  | Labels | Precision | Recall | F1 score | AUC ROC | Accuracy | Specificity | support |
|---|---|---|---|---|---|---|---|---|
|  | CAP | 0.9958 | 0.9728 | 0.9842 | 0.9989 | 0.9896 | 0.9979 | 3384 |
|  | Normal | 0.9745 | 0.9947 | 0.9845 | 0.9993 | 0.9896 | 0.9870 | 3384 |
|  | COVID | 0.9917 | 0.9941 | 0.9929 | 0.9996 | 0.9953 | 0.9959 | 3384 |
| **Micro avg** |  | 0.9872 | 0.9872 | 0.9872 | 0.9990 |  |  | 10152 |
| **Macro avg** |  | 0.9873 | 0.9872 | 0.9872 | 0.9992 |  |  | 10152 |
| **Weighted avg** |  | 0.9873 | 0.9872 | 0.9872 |  |  |  | 10152 |
| **Samples avg** |  | 0.9872 | 0.9872 | 0.9872 |  |  |  | 10152 |

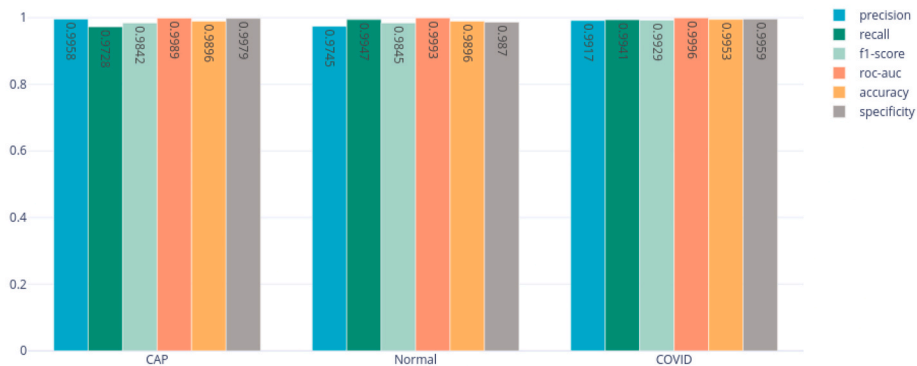

**Fig. 9.** Graphical representation of the class-wise score.

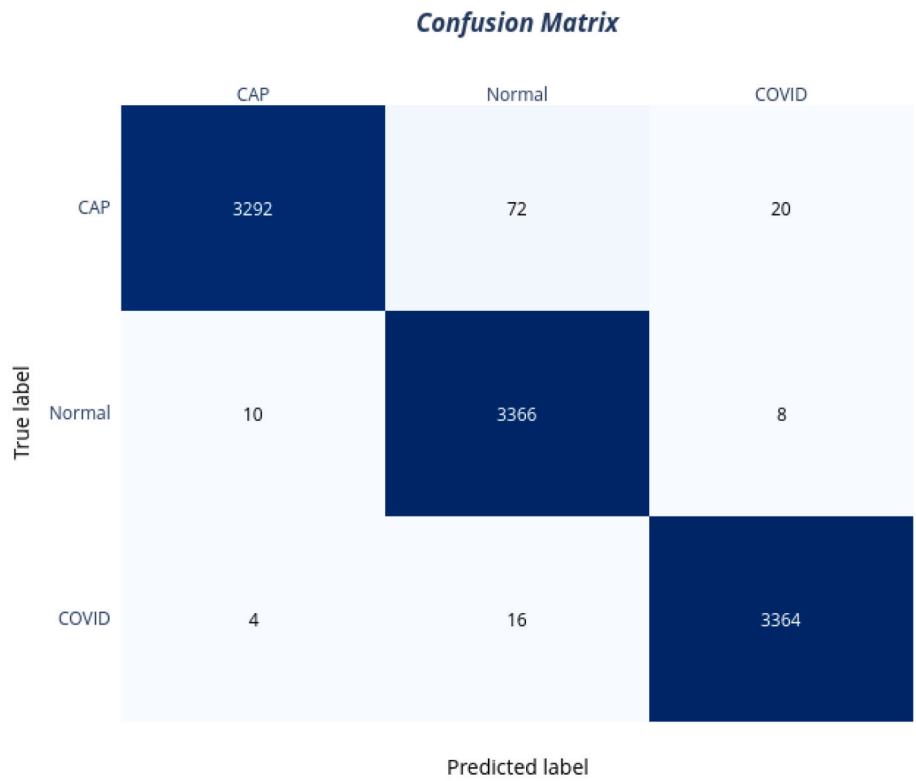

**Fig. 10.** Confusion matrix of the CovidExpert.

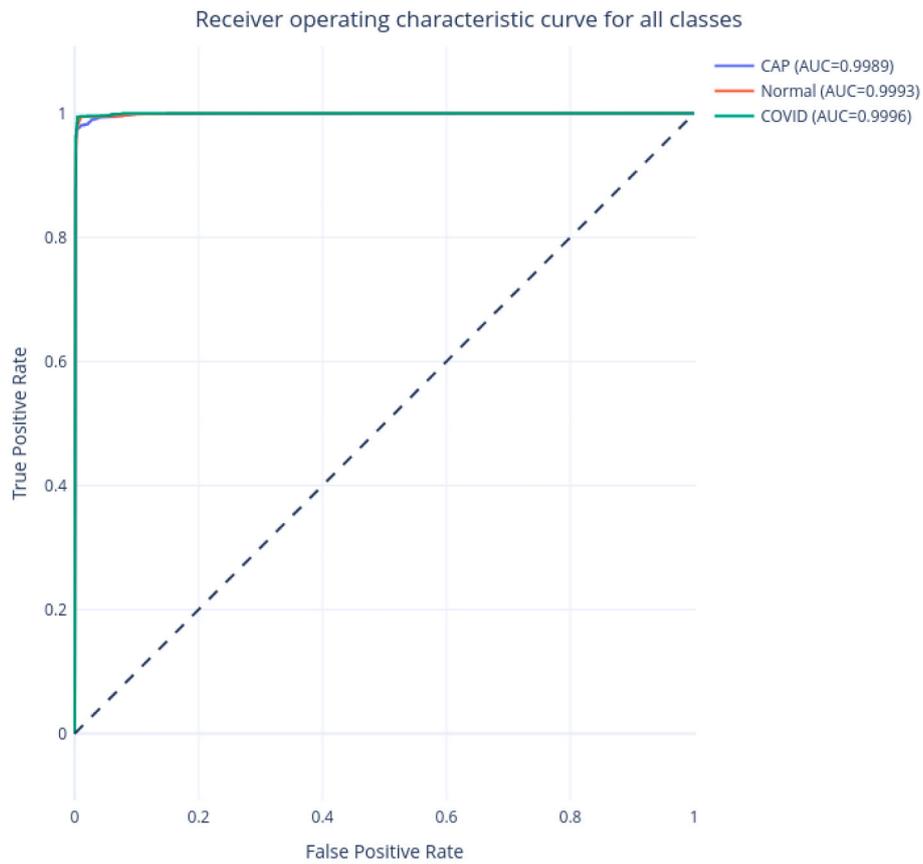

**Fig. 11.** ROC curve of the proposed system.

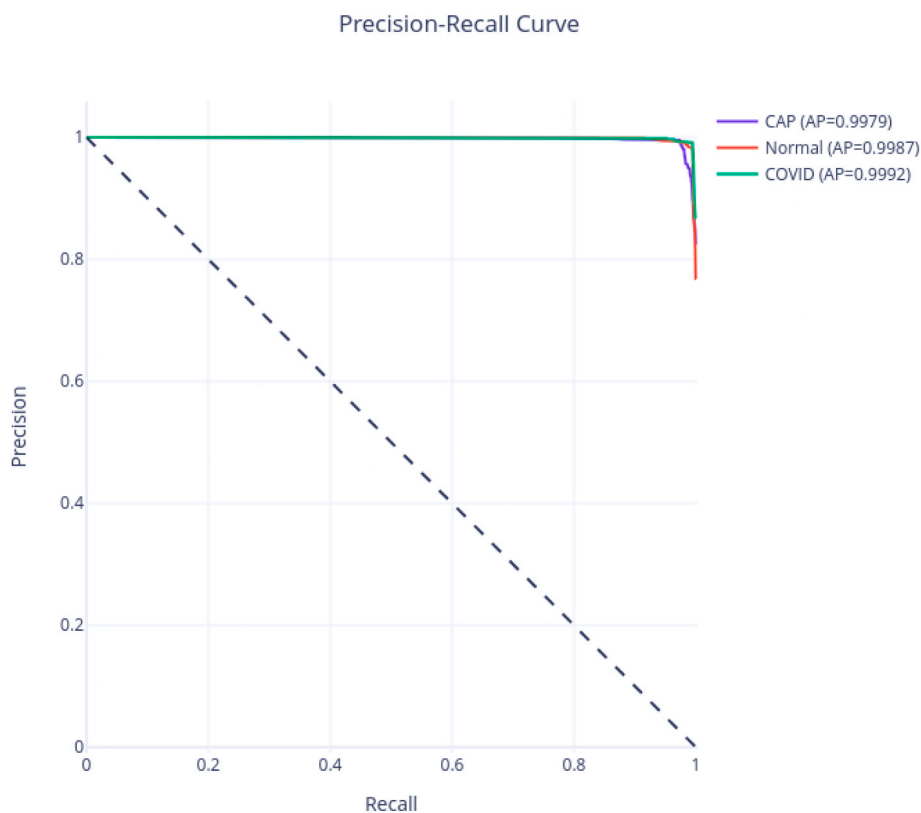

**Fig. 12.** PR curve of the proposed system.

**Table 7**
Comparison of the proposed system with existing systems.

| Author | Architecture | Score, Metric | Classification | Dataset | Data type |
| --- | --- | --- | --- | --- | --- |
| Panwar et al. [13] | VGG-19 | 95.61%, accuracy | Binary | (Training, testing) = (6090, 1522) | images |
| Horry et al. [18] | VGG-19 | 84.0%, accuracy | Binary | (Training, testing) = (746, 146) | |
| Rahimzadeh et al. [14] | ResNet50V2 | 98.49%, accuracy | Binary | (Training, testing) = (55853, 7996) | |
| Hussain et al. [15] | CoroDet | 99.1%, accuracy (2 classes) | Binary & Multi-class | For 2 classes: (Training, testing) = (1300, 260) | |
| | | 94.2%, accuracy (3 classes) | | For 3 classes: (Training, testing) = (2100, 420) | |
| | | 91.2%, accuracy (4 classes) | | For 4 classes: (Training, testing) = (2100, 420) | |
| Zhao et al. [19] | Bit-M | 99.2%, accuracy | Binary | (Training, testing) = (194922, 25658) | |
| Serte et al. [20] | ResNet-50 | 96%, AUC | Binary | (Training, Testing) = (244, 75) | 3D CT scans |
| Mukherjee et al. [21] | Tailored CNN | 96.28%, accuracy | Binary | (Training, Testing) = (672, 672) | images |
| Shah et al. [22] | VGG-19 | 94.5%, accuracy | Binary | (Training, Testing) = (592, 73) | |
| Maftouni et al. [23] | DenseNet-121+ResAttNet-92 | 95.31%, accuracy | Multi-class (3 classes) | (Training, Testing) = (14385, 1238) | |
| Shorfuzzaman et al. [24] | Siamese Network (VGG-16 encoder) | 95.6%, accuracy | Multi-class (3 classes) | (Training, Testing) = (30, 648) | |
| Jadon et al. [25] | Siamese Network (VGG-16 encoder) | 96.4%, accuracy | Multi-class (3 classes) | (Training, Testing) = (2520, 840) | |
| Kogilavani et al. [16] | VGG16 | 97.68%, accuracy | Binary | (Training, Testing) = (2711, 581) | |
| Basu et al. [17] | CNN +KNN | 98.87%, accuracy | Binary | (Training, Testing) = (2487, 439) | |
| **Proposed System** | **Triplet Siamese Network (Ensemble of CNNs as base encoder)** | **98.719%, accuracy** | **Multi-class (3 classes)** | **(Training, Testing) = (600, 10152)** | |

recognize every positive case during this health emergency. This work proposed a cutting-edge AI-based approach for an urgently needed quick and accurate diagnostic method for COVID-19 disease. The proposed research aims to achieve this by integrating a few-shot learning model with the similarity learning method and an ensemble of pre-trained CNN encoders. With a small number of training data, we demonstrated the value of few-shot learning for multi-class classification by automatically classifying COVID-19, CAP, and Normal cases from CT scans to expedite patient care. We presented a novel architecture that combines few-shot learning with an ensemble of pre-trained convolutional neural networks to retrieve feature vectors from images which are then fed into the Triplet Siamese Network to discover similarities between input images. The ROC score was 99.9%, the specificity was 99.36%, the sensitivity was 98.72%, and the overall accuracy of the suggested model for the multi-class classification was 98.719%. These performance values are thought to be very important for applications in medical contexts. Our suggested model performs similarly or better than the currently published models. This is encouraging since only 600 training samples were used to train our few-shot learning model. We hope that the shortcomings of CNNs may be overcome by using our architecture on a small and unbalanced dataset. Our strategy may facilitate the work of radiologists. Our method has shown a bright future in providing radiologists and clinicians with second views. The suggested system has certain drawbacks. Since Siamese neural networks (SNN) use quadratic pairs or cubic functions to learn from, they often need longer training time than conventional neural networks (pointwise learning). Because pairwise learning is a component of SNN training, the output is the distance from each class or similarity rather than the prediction probabilities. The suggested model will not work in other CT scan views, such as sagittal and coronal, since the dataset only included the axial view of abdominal CT scans. The research used imaging data; it did not consider the patient's age, gender, or past medical history. We want to expand our effort by resolving these issues soon. Developing the suggested method, such as reducing the training dataset and locating the region of irregularities, is one of our main research priorities for the future.

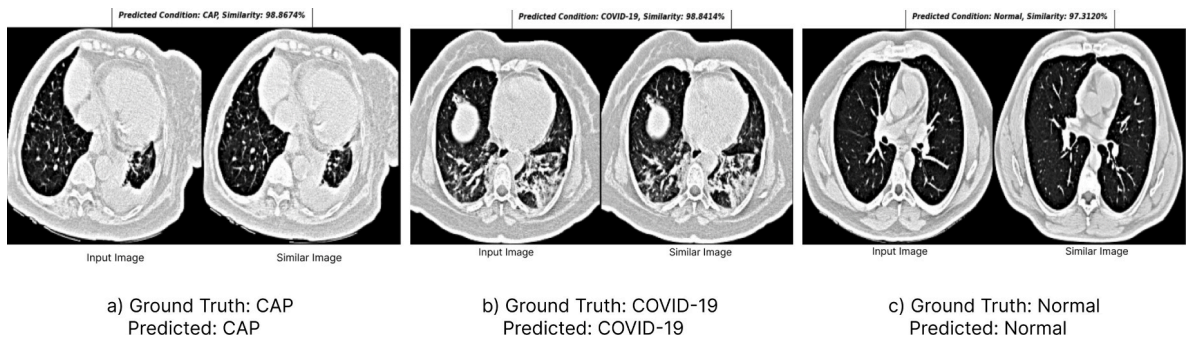

Fig. 13. Correct predictions.

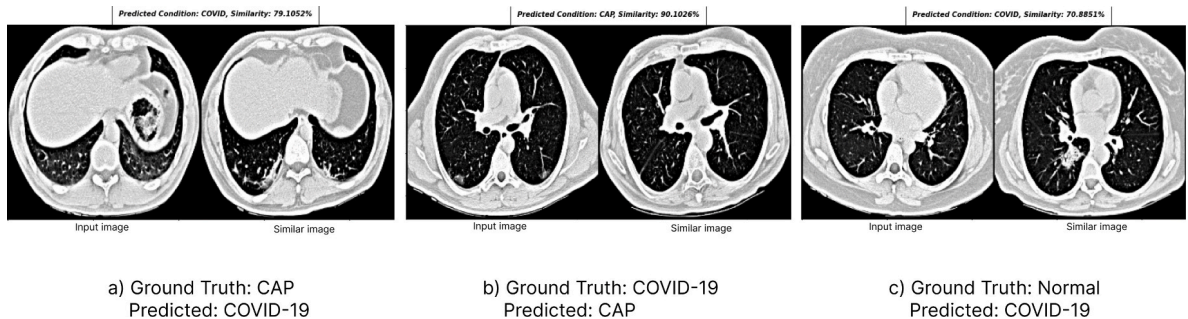

Fig. 14. Wrong predictions.

**Declaration of competing interest**

The authors declare that they have no known competing financial interests or personal relationships that could have appeared to influence the work reported in this paper.

**Acknowledgements**

The authors would like to thank Syed Md. Shakawath Hossain, Medical Technologist, CT Scan department, Medinova Medical Services Ltd, for his assistance in creating the dataset for the meta-learning model. The authors wish to acknowledge the Department of Computer Science and Engineering (CSE), Shahjalal University of Science & Technology (SUST) for providing technical support.

**References**

[1] COVID Live. Coronavirus Statistics - worldometer [internet]. n.d. "Worldometers. info. 2022. https://www.worldometers.info/coronavirus/.
[2] Abebe Endeshaw Chekol, Dejenie Tadesse Asmamaw, Shiferaw Mestet Yibeltal, Malik Tabarak. The newly emerged COVID-19 disease: a systemic review. Virol J 2020;17(1):1–8. https://doi.org/10.1186/s12985-020-01363-5.
[3] Wang Shuai, Kang Bo, Ma Jinlu, Zeng Xianjun, Xiao Mingming, Guo Jia, Cai Mengjiao, et al. A deep learning algorithm using CT images to screen for corona virus disease (COVID-19). Eur Radiol 2021;31(8):6096–104. https://doi.org/10.1007/s00330-021-07715-1.
[4] Fang Yicheng, Zhang Huangqi, Xie Jicheng, Lin Minjie, Ying Lingjun, Pang Peipei, Ji Wenbin. Sensitivity of chest CT for COVID-19: comparison to RT-PCR. Radiology 2020. https://doi.org/10.1148/radiol.2020200432.
[5] Emon Md Moniruzzaman, Rahman Ornob Tareque, Rahman Moqsadur. Predicting skull fractures via CNN with classification algorithms. In: Proceedings of the 2nd international conference on computing advancements; 2022. p. 442–9. https://doi.org/10.1145/3542954.3543017.
[6] Emon Md Moniruzzaman, Rahman Ornob Tareque, Rahman Moqsadur. Classifications of skull fractures using CT scan images via CNN with lazy learning approach. J Comput Sci 2022;18(3):116–29. https://doi.org/10.3844/jcssp.2022.116.129.
[7] Islam Nayaar, Ebrahimzadeh Sanam, Salameh Jean-Paul, Kazi Sakib, Fabiano Nicholas, Lee Treanor, Absi Marissa, et al. Thoracic imaging tests for the diagnosis of COVID-19. Cochrane Database Syst Rev 2021;(3). https://doi.org/10.1002/14651858.CD013639.pub2.
[8] Zhao Amy, Guha Balakrishnan, Durand Fredo, Guttag John V, V Dalca Adrian. Data augmentation using learned transformations for one-shot medical image segmentation. In: Proceedings of the IEEE/CVF conference on computer vision and pattern recognition; 2019. https://doi.org/10.48550/arXiv.1902.09383. 8543–53.
[9] Thrun S, Pratt L. Learning to learn. Springer Science & Business Media; 2012. https://doi.org/10.1007/978-1-4615-5529-2.
[10] Deng Shumin, Zhang Ningyu, Kang Jiaojian, Zhang Yichi, Zhang Wei, Chen Huajun. Meta-learning with dynamic-memory-based prototypical network for few-shot event detection. In: Proceedings of the 13th international conference on web search and data mining; 2020. https://doi.org/10.1145/3336191.3371796. 151–59.
[11] Reeder John. Visualizing features from deep neural networks trained on alzheimer's disease and few-shot learning models for alzheimer's disease. 2021.
[12] Dey Sounak, Dutta Anjan, Toledo J Ignacio, Ghosh Suman K, Lladós Josep, Pal Umapada. Signet: convolutional siamese network for writer independent offline signature verification. 2017. https://doi.org/10.48550/arXiv.1707.02131. *arXiv Preprint arXiv:1707.02131*.
[13] Panwar Harsh, Gupta PK, Siddiqui Mohammad Khubeb, Morales-Menendez Ruben, Bhardwaj Prakhar, Singh Vaishnavi. A deep learning and grad-CAM based color visualization approach for fast detection of COVID-19 cases using chest x-ray and

CT-scan images. Chaos, Solit Fractals 2020;140:110190. https://doi.org/10.1016/j.chaos.2020.110190.
[14] Rahimzadeh Mohammad, Attar Abolfazl, Sakhaei Seyed Mohammad. A fully automated deep learning-based network for detecting covid-19 from a new and large lung ct scan dataset. Biomed Signal Process Control 2021;68:102588. https://doi.org/10.1016/j.bspc.2021.102588.
[15] Hussain Emtiaz, Hasan Mahmudul, Rahman Md Anisur, Lee Ickjai, Tamanna Tasmi, Parvez Mohammad Zavid. CoroDet: a deep learning based classification for COVID-19 detection using chest x-ray images. Chaos, Solit Fractals 2021;142:110495. https://doi.org/10.1016/j.chaos.2020.110495.
[16] Kogilavani SV, Prabhu J, Sandhiya R, Sandeep Kumar M, Subramaniam UmaShankar, Karthick Alagar, Muhibbullah M, Imam Sharmila Banu Sheik. COVID-19 detection based on lung CT scan using deep learning techniques. Comput Math Methods Med 2022. https://doi.org/10.1155/2022/7672196. 2022.
[17] Basu Arpan, Hassan Sheikh Khalid, Cuevas Erik, Sarkar Ram. COVID-19 detection from CT scans using a two-stage framework. Expert Syst Appl 2022;193:116377. https://doi.org/10.1016/j.eswa.2021.116377.
[18] Horry Michael J, Chakraborty Subrata, Paul Manoranjan, Ulhaq Anwaar, Pradhan Biswajeet, Saha Manas, Shukla Nagesh. COVID-19 detection through transfer learning using multimodal imaging data. IEEE Access 2020;8. https://doi.org/10.1109/ACCESS.2020.3016780. 149808–24.
[19] Zhao Wentao, Jiang Wei, Qiu Xinguo. Deep learning for COVID-19 detection based on CT images. Sci Rep 2021;11(1):1–12. https://doi.org/10.1038/s41598-021-93832-2.
[20] Serte Sertan, Demirel Hasan. Deep learning for diagnosis of COVID-19 using 3D CT scans. Comput Biol Med 2021;132:104306. https://doi.org/10.1016/j.compbiomed.2021.104306.
[21] Mukherjee Himadri, Ghosh Subhankar, Dhar Ankita, Obaidullah Sk Md, Santosh KC, Roy Kaushik. Deep neural network to detect COVID-19: one architecture for both CT scans and chest x-rays. Appl Intell 2021;51(5):2777–89. https://doi.org/10.1007/s10489-020-01943-6.
[22] Shah Vruddhi, Keniya Rinkal, Shridharani Akanksha, Punjabi Manav, Shah Jainam, Mehendale Ninad. Diagnosis of COVID-19 using CT scan images and deep learning techniques. Emerg Radiol 2021;28(3):497–505. https://doi.org/10.1007/s10140-020-01886-y.
[23] Maftouni Maede, Law Andrew Chung Chee, Shen Bo, Grado Zhenyu James Kong, Zhou Yangze, Yazdi Niloofar Ayoobi. A robust ensemble-deep learning model for COVID-19 diagnosis based on an integrated CT scan images database. In: IIE annual conference. Proceedings, vols. 632–37. Institute of Industrial; Systems Engineers (IISE); 2021.
[24] Shorfuzzaman Mohammad, Hossain M Shamim. MetaCOVID: a siamese neural network framework with contrastive loss for n-shot diagnosis of COVID-19 patients. Pattern Recogn 2021;113:107700. https://doi.org/10.1016/j.patcog.2020.107700.
[25] Jadon Shruti. COVID-19 detection from scarce chest x-ray image data using few-shot deep learning approach. In: Medical imaging 2021: imaging informatics for healthcare, *research, and applications*. SPIE; 2021. https://doi.org/10.1117/12.2581496. 11601:161–70.
[26] Deng Jia, Dong Wei, Socher Richard, Li Li-Jia, Li Kai, Fei-Fei Li. Imagenet: a large-scale hierarchical image database. In: 2009 *IEEE Conference on computer Vision and pattern recognition*. Ieee; 2009. https://doi.org/10.1109/CVPR.2009.5206848. 248–55.
[27] Krizhevsky Alex, Sutskever Ilya, Hinton Geoffrey E. Imagenet classification with deep convolutional neural networks. Commun ACM 2017;60(6):84–90. https://doi.org/10.1145/3065386.
[28] Szegedy Christian, Liu Wei, Jia Yangqing, Sermanet Pierre, Scott Reed, Anguelov Dragomir, Erhan Dumitru, Vincent Vanhoucke, Rabinovich Andrew. Going deeper with convolutions. In: Proceedings of the IEEE conference on computer vision and pattern recognition; 2015. p. 1–9. https://doi.org/10.1109/CVPR.2015.7298594.
[29] Howard Andrew, Sandler Mark, Chu Grace, Chen Liang-Chieh, Chen Bo, Tan Mingxing, Wang Weijun, et al. Searching for Mobilenetv3. In: Proceedings of the IEEE/CVF international conference on computer vision; 2019. https://doi.org/10.1109/ICCV.2019.00140. 1314–24.
[30] Dosovitskiy Alexey, Beyer Lucas, Alexander Kolesnikov, Weissenborn Dirk, Zhai Xiaohua, Unterthiner Thomas, Dehghani Mostafa, et al. An image is worth 16x16 words: transformers for image recognition at scale. 2020. https://doi.org/10.48550/arXiv.2010.11929. *arXiv Preprint arXiv:2010.11929*.
[31] Liu Ze, Lin Yutong, Cao Yue, Han Hu, Wei Yixuan, Zhang Zheng, Lin Stephen, Guo Baining. Swin transformer: hierarchical vision transformer using shifted windows. In: Proceedings of the IEEE/CVF international conference on computer vision; 2021. https://doi.org/10.48550/arXiv.2103.14030. 10012–22.
[32] Kolesnikov Alexander, Beyer Lucas, Zhai Xiaohua, Puigcerver Joan, Jessica Yung, Gelly Sylvain, Houlsby Neil. Big transfer (bit): general visual representation learning. In: European conference on computer vision, vols. 491–507. Springer; 2020. https://doi.org/10.1007/978-3-030-58558-7_29.
[33] Huang Gao, Liu Zhuang, Van Der Maaten Laurens, Weinberger Kilian Q. Densely connected convolutional networks. In: Proceedings of the IEEE conference on computer vision and pattern recognition; 2017. p. 4700–8. https://doi.org/10.48550/arXiv.1608.06993.
[34] Sandler Mark, Howard Andrew, Zhu Menglong, Zhmoginov Andrey, Chen Liang-Chieh. Mobilenetv2: inverted residuals and linear bottlenecks. In: Proceedings of the IEEE conference on computer vision and pattern recognition; 2018. p. 4510–20. https://doi.org/10.48550/arXiv.1801.04381.
[35] Tan Mingxing, Le Quoc. Efficientnet: rethinking model scaling for convolutional neural networks. In: *International Conference on machine learning*, 6105–14. PMLR; 2019. https://doi.org/10.48550/arXiv.1905.11946.
[36] Xie Saining, Girshick Ross, Dollár Piotr, Tu Zhuowen, He Kaiming. Aggregated residual transformations for deep neural networks. In: Proceedings of the IEEE conference on computer vision and pattern recognition; 2017. p. 1492–500. https://doi.org/10.48550/arXiv.1611.05431.
[37] Lin Tsung-Yi, Goyal Priya, Girshick Ross, He Kaiming, Dollár Piotr. Focal loss for dense object detection. In: Proceedings of the IEEE international conference on computer vision; 2017. https://doi.org/10.48550/arXiv.1708.02002. 2980–88.